%% file: MAIN.tex
\documentclass[sigconf]{acmart}

\AtBeginDocument{%
  \providecommand\BibTeX{{%
    \normalfont B\kern-0.5em{\scshape i\kern-0.25em b}\kern-0.8em\TeX}}}

\copyrightyear{2021}
\acmYear{2021}
\setcopyright{acmcopyright}
\acmConference[WSDM '21] {Proceedings of the Fourteenth ACM International Conference on Web Search and Data Mining}{March 8--12, 2021}{Virtual Event, Israel}
\acmBooktitle{Proceedings of the Fourteenth ACM International Conference on Web Search and Data Mining (WSDM '21), March 8--12, 2021, Virtual Event, Israel}
\acmPrice{15.00}
\acmDOI{10.1145/3437963.3441735}
\acmISBN{978-1-4503-8297-7/21/03}

\usepackage{subfig}
\usepackage{amsmath,enumitem}
\usepackage[ruled,vlined,linesnumbered,noresetcount]{algorithm2e}
\usepackage{threeparttable,color}
\usepackage{booktabs}
\usepackage{multirow}
\usepackage{balance}

\begin{document}
\fancyhead{}
\title{Node Similarity Preserving Graph Convolutional Networks}


\author{Wei Jin}
\affiliation{%
  \institution{Michigan State University}
}
\email{jinwei2@msu.edu}

\author{Tyler Derr}
\affiliation{%
  \institution{Vanderbilt University}
}
\email{tyler.derr@vanderbilt.edu}

\author{Yiqi Wang}
\affiliation{%
  \institution{Michigan State University}
}
\email{wangy206@msu.edu}

\author{Yao Ma}
\affiliation{%
  \institution{Michigan State University}
}
\email{mayao4@msu.edu}

\author{Zitao Liu}
\authornote{The corresponding author: Zitao Liu.}
\affiliation{%
  \institution{TAL Education Group}
}
\email{liuzitao@100tal.com}

\author{Jiliang Tang}
\affiliation{%
  \institution{Michigan State University}
}
\email{tangjili@msu.edu}


\newcommand{\modelname}{Pro-GNN }
\newcommand{\SVDpaper}{GCN-SVD }
\newcommand{\metattack}{\textit{metattack }}
\newcommand{\nettack}{\textit{\textit{nettack }}}

\begin{abstract}
\input{section/abstract.tex}

\end{abstract}

\maketitle

\input{section/intro}

\input{section/relatedv2.tex}

\input{section/preliminary.tex}
\input{section/frameworkv2}

\input{section/experiment.tex}

\input{section/conclusion.tex}

\balance 
\bibliographystyle{ACM-Reference-Format}
\bibliography{sample}


\end{document}

%% file: section/abstract.tex
Graph Neural Networks (GNNs) have achieved tremendous success in various real-world applications due to their strong ability in graph representation learning. GNNs explore the graph structure and node features by aggregating and transforming information within node neighborhoods. However, through theoretical and empirical analysis, we reveal that the aggregation process of GNNs tends to destroy node similarity in the original feature space. There are many scenarios where node similarity plays a crucial role. Thus, it has motivated the proposed framework SimP-GCN that can effectively and efficiently preserve node similarity while exploiting graph structure. Specifically, to balance information from graph structure and node features, we propose a feature similarity preserving aggregation which adaptively integrates graph structure and node features. Furthermore, we employ self-supervised learning to explicitly capture the complex feature similarity and dissimilarity relations between nodes. We validate the effectiveness of SimP-GCN on seven benchmark datasets including three assortative and four disassorative graphs. The results demonstrate that SimP-GCN outperforms representative baselines. Further probe shows various advantages of the proposed framework. The implementation of SimP-GCN is available at \url{https://github.com/ChandlerBang/SimP-GCN}.

%% file: section/intro.tex
\section{Introduction}
\label{sec:intro}

Graphs are essential data structures that describe pairwise relations between entities for real-world data from numerous domains such as social media, transportation, linguistics and chemistry~\cite{ma2020deep,battaglia2018relational-survey,wu2019comprehensive-survey,zhou2018graph-survey}.  Many important tasks on graphs involve predictions over nodes and edges. For example, in node classification,  we aim to predict labels of unlabeled nodes~\cite{kipf2016semi,gat, li2017effective}; and 
in link prediction, we want to infer whether there exists an edge between a given node pair~\cite{vgae,zhang2018link}. All these tasks can be tremendously facilitated if good node representations can be extracted. 

Recent years have witnessed remarkable achievements in Graph Neural Networks (GNNs) for learning node  representations that have advanced various tasks on graphs~\cite{ma2020deep,zhou2018graph-survey,wu2019comprehensive-survey,battaglia2018relational-survey}. GNNs explore the graph structure and node features by aggregating and transforming information within node neighborhoods. During the aggregation process, GNNs tend to smooth the node features. Note that in this work, we focus on one of the most popular GNN models, i.e., graph convolution network (or GCN)~\cite{kipf2016semi}; however, our study can be easily extended to other GNN models as long as they follow a neighborhood aggregation process. By connecting GCN with Laplacian smoothing, we show that the graph convolution operation in GCN is equivalent to one-step optimization of the signal recovering problem (see Section~\ref{sec:lapsmooth}). This operation will reduce the overall node feature difference. Furthermore, through empirical study, we demonstrate that (1) given a node, nodes that have top feature similarity with the node are less likely to overlap with these that connect to the node ( we refer to this as the `` Less-Overlapping" problem in this work); and (2) GCN tends to preserve structural similarity rather than node similarity, no matter whether the graph is assortative (when node homophily~\cite{mcpherson2001birds} holds in the graph) or not (see Section~\ref{sec:overlapping}).

Neighbor smoothing from the aggregation process and the less-overlapping problem inevitably introduce one consequence -- node representations learned by GCN naturally destroy the node similarity of the original feature space. This consequence can reduce the effectiveness of the learned representations and hinder the performance of downstream tasks since node similarity can play a crucial role in many scenarios. Next we illustrate several examples of these scenarios. First, when the graph structure is not optimal, e.g., graphs are disassortative~\cite{newman2002assortative} or graph structures have been manipulated by adversarial attacks, relying heavily on the structure information can lead to very poor performance~\cite{pei2020geom,nettack,jin2020adversarial,xu2019adversarial}. Second, since low-degree nodes only have a small number of neighbors, they receive very limited information from neighborhood and GNNs often cannot perform well on those nodes~\cite{tang2020graph-degree}.  This scenario is similar to the cold-start problem in recommender systems where the system cannot provide a satisfying recommendation for new users or items~\cite{bobadilla2013recommender}. Third, for nodes affiliated to the ``hub" nodes which presumably connect with many different types of nodes, the information from the hub nodes could be very confusing and misleading~\cite{xu2018representation}. Fourth, when applying multiple graph convolution layers, GNNs can suffer over-smoothing problem that the learned node embeddings become totally indistinguishable~\cite{li2018deeper-oversmooth,liu2020towards-oversmooth,rong2019dropedge}. 

In this work, we aim to design a new graph convolution model that can better preserve the original node similarity. In essence, we are faced with two challenges. First, \textbf{ how to balance the information from graph structure and node features during aggregation?} We propose an adaptive strategy that coherently integrates the graph structure and node features in a data-driven way. It enables each node to adaptively adjust the information from graph structure and node features. Second, \textbf{how to explicitly capture the complex pairwise feature similarity relations?}  We employ a self-supervised learning strategy to predict the pairwise feature similarity from the hidden representations of given node pairs. By adopting similarity prediction as the self-supervised pretext task, we are allowed to explicitly encode the pairwise feature relation. Furthermore, combining the above two components leads to our proposed model, SimP-GCN, which can effectively and adaptively preserve feature and structural similarity, thus achieving state-of-the-art performance on a wide range of benchmark datasets. Our contributions can be summarized as follows:
\begin{itemize}
    \item We theoretically and empirically show that GCN can destroy original node feature similarity.
    \item We propose a novel GCN model that effectively preserves feature similarity and adaptively balances the information from graph structure and node features while simultaneously leveraging their rich information.
    \item Extensive experiments have demonstrated that the proposed framework can outperform representative baselines on both assortative and disassortative graphs. We also show that preserving feature similarity can significantly boost the robustness of GCN against adversarial attacks.
\end{itemize}


%% file: section/relatedv2.tex
\section{Related Work}
Over the past few years, increasing efforts have been devoted toward generalizing deep learning to graph structured data in the form of graph neural networks. There are mainly two streams of graph neural networks, i.e. spectral based methods and spatial based methods~\cite{ma2020deep}. Spectral based methods learn node representations based on graph spectral theory~\cite{shuman2013emerging}. Bruna et al.~\cite{bruna2013spectral} first generalize convolution operation to non-grid structures from spectral domain 
by using the graph Laplacian matrix. Following this work, ChebNet~\cite{ChebNet} utilizes Chebyshev polynomials to modulate the graph Fourier coefficients and simplify the convolution operation. The ChebNet is further simplified to GCN~\cite{kipf2016semi} by setting the order of the polynomial to $1$ together with other approximations. While being a simplified spectral method, GCN can be also regarded as a spatial based method. From a node perspective, when updating its node representation, GCN aggregates information from its neighbors. Recently, many more spatial based methods with different designs to aggregate and transform the neighborhood information are proposed including GraphSAGE~\cite{graphsage}, MPNN~\cite{mpnn}, GAT~\cite{gat}, etc.

While graph neural networks have been demonstrated to be effective in many applications, their performance might be impaired when the graph structure is not optimal. For example, their performance deteriorates greatly on disassortative graphs where homophily does not hold~\cite{pei2020geom} and thus the graph structure introduces noises; adversarial attack can inject carefully-crafted noise to disturb the graph structure and fool graph neural networks into making wrong predictions~\cite{nettack,mettack,wu2019adversarial-jaccard}. In such cases, the graph structure information may not be optimal for graph neural networks to achieve better performance while the original features could come as a rescue if carefully utilized.  Hence, it urges us to develop a new graph neural network capable of preserving node similarity in the original feature space. 


%% file: section/preliminary.tex
\section{Preliminary Study}\label{sec:prelimstudy}
In this section, we investigate if the GCN model can preserve feature similarity via theoretical analysis and empirical study. Before that, we first introduce key notations and concepts. 

Graph convolutional network (GCN)~\cite{kipf2016semi} was proposed to solve semi-supervised node classification problem, where only a subset of the nodes with labels. A graph is defined as $\mathcal{G}=({\bf\mathcal{V}},{\bf\mathcal{E}}, {\bf X})$, where  $\mathcal{V}=\{v_1, v_2, ..., v_n\}$ is the set of $n$ nodes, $\mathcal{E}$ is the set of edges describing the relations between nodes and ${\bf X} = [{\bf x}^{\top}_1,{\bf x}^\top_2,\ldots,{\bf x}^\top_n]\in\mathbb{R}^{n\times{d}}$ is the node feature matrix where $d$ is the number of features and ${\bf x}_i$ indicates the node features of $v_i$. The graph structure can also be represented by an adjacency matrix $\mathbf{A} \in \{0,1\}^{n \times n}$ where $\mathbf{A}_{ij}=1$ indicates the existence of an edge between nodes $v_i$ and $v_j$, otherwise $\mathbf{A}_{ij}=0$. A single graph convolutional filter with parameter $\theta$ takes the adjacency matrix ${\bf A}$ and a graph signal ${\bf f}\in \mathbb{R}^{n}$ as input. It generates a filtered graph signal ${\bf f}^{\prime}$ as:
\begin{equation}
{\bf f}^{\prime}= \theta\thinspace{\bf \tilde{ D}}^{-\frac{1}{2}}{\bf \tilde{A}}{\bf \tilde{D}}^{-\frac{1}{2}}\thinspace {\bf f},
\label{eq:gc}
\end{equation}
where ${\bf \tilde{A}} = {\bf A} + {\bf I}$ and $\tilde{\bf D}$ is the diagonal matrix of ${\bf \tilde{A}}$ with $\tilde{\bf D}_{ii} = \sum_{j} \tilde{\bf A}_{ij}$. The node representations of all nodes can be viewed as multi-channel graph signals and the $l$-th graph convolutional layer with non-linearity is rewritten in a matrix form as:
\begin{equation}
    {\bf H}^{(l)} = \sigma(\tilde{\bf D}^{-1 / 2}{\bf \tilde{A}} \tilde{\bf D}^{-1 / 2}{\bf H}^{(l-1)}{\bf W}^{(l)}),
\end{equation}
where ${\bf H}^{(l)}$ is the output of the $l$-th layer, ${\bf H}^{(0)}={\bf X}$, ${\bf W}^{(l)}$ is the weight matrix of the $l$-th layer and $\operatorname{\sigma}$ is the ReLU activation function.

\subsection{Laplacian Smoothing in GCN}
\label{sec:lapsmooth}
As shown in Eq.~(\ref{eq:gc}), GCN naturally smooths the features in the neighborhoods as each node aggregates information from neighbors. This characteristic is related to Laplacian smoothing and it essentially destroys node similarity in the original feature space. While Laplacian smoothing in GCN has been studied~\cite{li2018deeper-oversmooth,yang2020revisiting-oversmooth}, in this work we revisit this process from a new perspective.

Given a graph signal $\bf f$ defined on a graph $\mathcal{G}$ with normalized Laplacian matrix ${\bf {L}} = {\bf I} -  {\bf \tilde{ D}}^{-\frac{1}{2}}{\bf \tilde{A}}{\bf \tilde{D}}^{-\frac{1}{2}}$, the signal smoothness over the graph can be calculated as:
\begin{equation}
{\bf f}^{T} {\bf {L}} {\bf f}=\frac{1}{2} \sum_{i,j} {\bf \tilde{A}}_{i j}\left(\frac{{\bf f}_{i}}{\sqrt{1+d_{i}}}-\frac{{\bf f}_{j}}{\sqrt{1+d_{j}}}\right)^{2}.
\end{equation}
where $d_i$ and $d_j$ denotes the degree of nodes $v_i$ and $v_j$ respectively. Thus a smaller value of ${\bf f}^{T} {\bf L} {\bf f}$ indicates smoother graph signal, i.e., smaller signal difference between adjacent nodes.

Suppose that we are given a noisy graph signal ${\bf f}_0 = {\bf f}^*+\eta$, where $\eta$ is uncorrelated additive Gaussian noise. The original signal ${\bf f}^*$ is assumed to be smooth with respect to the underlying graph $\mathcal{G}$. Hence, to recover the original signal, we can adopt the following objective function:
\begin{equation}
    \arg\min_{\bf f}~~ g({\bf f}) = \|{\bf f}-{\bf f}_0\|^2 + c{\bf f}^T {\bf L} {\bf f} ,
\label{eq:opti_signal}
\end{equation}
Then we have the following lemma.
\begin{lemma}
\label{lemma:gcn}
The GCN convolutional operator is one-step gradient descent optimization of Eq.~(\ref{eq:opti_signal}) when $c=\frac{1}{2}$.
\end{lemma}
\begin{proof}
The gradient of $g({\bf f})$ at ${\bf f}_0$ is calculated as:
\begin{equation}
    \nabla{g({\bf f}_0)} = 2({\bf f}_0-{\bf f}_0) + 2c{\bf L}{\bf f}_0 = 2c{\bf L}{\bf f}_0 .
\end{equation}
Then the one step gradient descent at ${\bf f}_0$ with learning rate $1$ is formulated as,
\begin{align}
{\bf f}_0 - \nabla{g({\bf f}_0)} & = {\bf f}_0 - 2c{\bf L}{\bf f}_0  \nonumber \\
& = ({\bf I} - 2c{\bf L}){\bf f}_0  \nonumber \\
& = ({\bf \tilde{ D}}^{-\frac{1}{2}}{\bf \tilde{A}}{\bf \tilde{D}}^{-\frac{1}{2}} + {\bf L} - 2c{\bf L}){\bf f}_0.
\end{align}
By setting $c$ to $\frac{1}{2}$, we finally arrives at,
\begin{equation}
   {\bf f}_0 - \nabla{g({\bf f}_0)} =  {\bf \tilde{ D}}^{-\frac{1}{2}}{\bf \tilde{A}}{\bf \tilde{D}}^{-\frac{1}{2}} {\bf f}_0. \nonumber 
\end{equation}
\vskip -1.5ex 
\end{proof}
By extending the above observation from the signal vector ${\bf f}$ to the feature matrix ${\bf X}$, we can easily connect the general form of the graph convolutional operator ${\bf \tilde{ D}}^{-\frac{1}{2}}{\bf \tilde{A}}{\bf \tilde{D}}^{-\frac{1}{2}}{\bf X}$ to Laplacian smoothing. Hence, the graph convolution layer tends to increase the feature similarity between the connected nodes. It is likely to destroy the original feature similarity. In particular, for those disassortative graphs where homphlily does not hold, this operation can bring in enormous noise.

\subsection{Empirical Study for GCN}
\label{sec:overlapping}
In the previous subsection, we demonstrated that GCN naturally smooths the features in the neighborhoods, consequently destroying the original feature similarity. In this subsection, we investigate if GCN can preserve node feature similarity via empirical study.

We perform our analysis on two assortative graph datasets (i.e., Cora and Citeseer) and two disassortative graph datasets (i.e., Actor and Cornell). The detailed statistics are shown in Table~\ref{tab:dataset}. For each dataset, we first construct two  $k$-nearest-neighbor ($k$NN) graphs based on original features and the hidden representations learned from GCN correspondingly. Together with the original graph, we have, in total, three graphs. Then we calculate the pairwise overlapping for these three graphs. We use ${\bf A}$, ${\bf A}_f$ and ${\bf A}_h$ to denote the original graph, the $k$NN graph constructed from original features and $k$NN graph from hidden representations, respectively. In this experiment, $k$ is set to $3$. Given a pair of adjacency matrices ${\bf A}_1$ and ${\bf A}_2$, we define the overlapping $OL({\bf A}_1, {\bf A}_2)$ between them as:  
\begin{equation}
    OL({\bf A}_1, {\bf A}_2) = \frac{\|{\bf A}_1 \cap {\bf A}_2\|}{\|{\bf A}_1\|},
\end{equation}
where ${\bf A}_1 \cap {\bf A}_2$ is the intersection of two matrices and $\|{\bf A}_1\|$ is the number of non-zero elements in ${\bf A_1}$. In fact, $OL({\bf A}_1, {\bf A}_2)$ indicates the overlapping percentage of edges in the two graphs ${\bf A}_1$ and ${\bf A}_2$. Larger value of $OL({\bf A}_1, {\bf A}_2)$ means larger overlapping between ${\bf A}_1$ and ${\bf A}_2$. We show the pairwise overlapping of these three graphs in Table~\ref{tab:overlap}. We make the following observations:

\begin{table}[h]
\caption{Overlapping percentage of different graph pairs.}
\vskip -0.25em
\begin{tabular}{@{}lcccc@{}}
\toprule
           Graph Pairs        & Cora    & Citeseer & Actor  & Cornell \\ \midrule
$OL({\bf A}_f, {\bf A}_h)$     & 3.15\%  & 3.73\%   & 1.15\% & 2.35\%  \\
$OL({\bf A}_h, {\bf A})$       & 21.24\% & 18.77\%  & 2.17\% & 7.74\% \\
$OL({\bf A}_f, {\bf A})$       & 3.88\%  & 3.78\%   & 0.03\% & 0.91\%  \\
\bottomrule
\end{tabular}
\label{tab:overlap}
\end{table}

\begin{itemize}[leftmargin=*]
    \item Assortative graphs (Cora and Citeseer) have a higher overlapping percentage of (${\bf A}_f, {\bf A}$) while in disassortative graphs (Actor and Cornell), it is much smaller. It indicates that in assortative graphs the structure information and feature information are highly correlated while in disassortative graphs they are not. It is in line with the definition of assortative and disassortative graphs.
    \item As indicated by $OL({\bf A}_f, {\bf A})$, the feature similarity is not very consistent with the graph structure. In assortative graphs, the overlapping percentage of (${\bf A}_f, {\bf A}_h$) is much lower than that of (${\bf A}_f, {\bf A}$). GCN will even amplify this inconsistency. In other words, although GCN takes both graph structure and node features as input, it fails to capture more feature similarity information than that the original adjacency matrix has carried.  
    \item By comparing the overlapping percentage of (${\bf A}_f, {\bf A}_h$) and (${\bf A}_h, {\bf A}$), we see the overlapping percentage of (${\bf A}_h, {\bf A}$) is consistently higher. Thus, GCN tends to preserve structure similarity instead of feature similarity regardless if the graph is asssortative or not.
\end{itemize}

%% file: section/frameworkv2.tex
\begin{figure*}[t]
    \centering
    \includegraphics[width=0.85\linewidth]{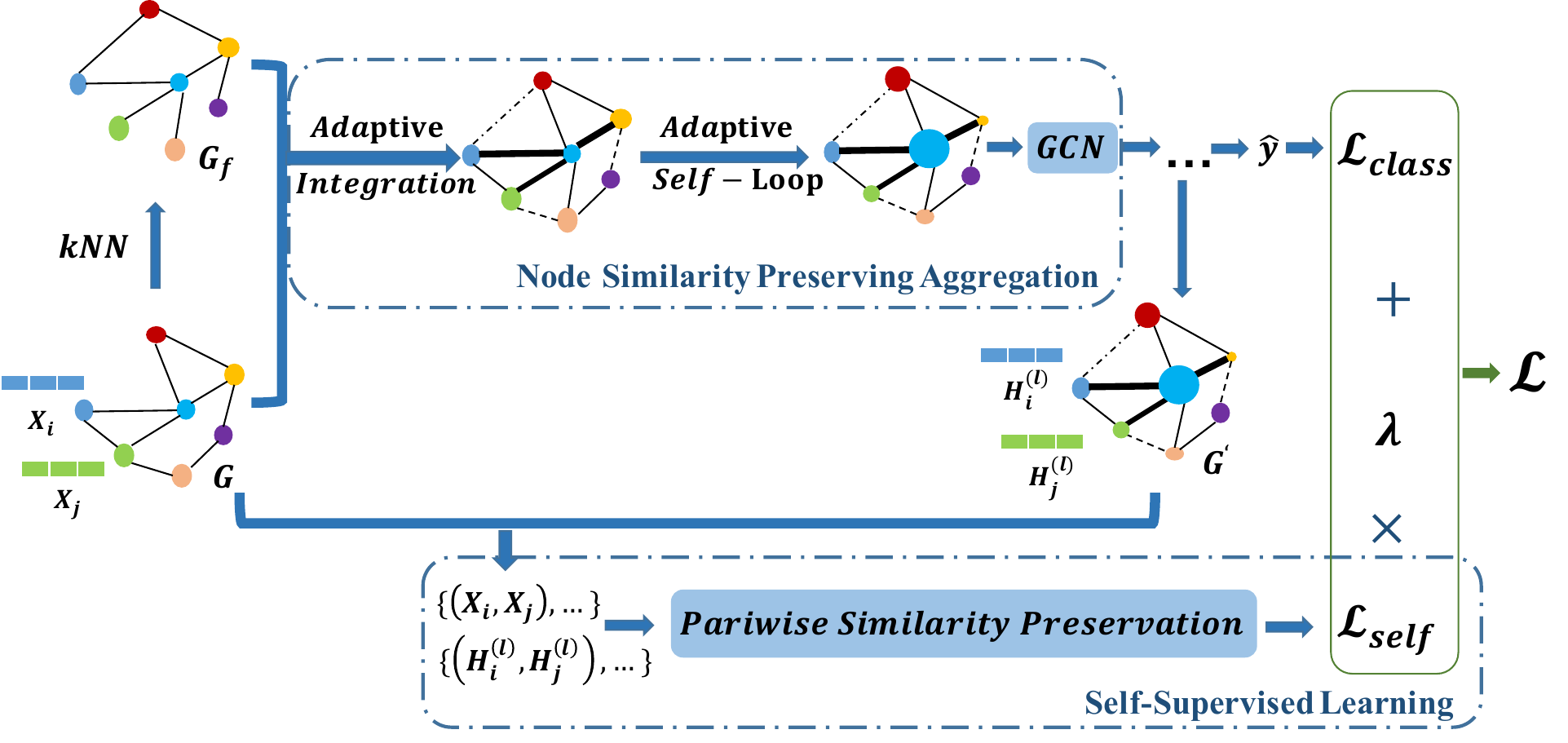}
    \vskip -0.5em
    \caption{An overall framework of the proposed SimP-GCN.}
    \label{fig:framework}
    \vskip -0.5em
\end{figure*}

\section{The Proposed Framework}

In this section, we design a node feature similarity preserving graph convolutional framework SimP-GCN by mainly solving the two challenges mentioned in Section~\ref{sec:intro}. An overview of SimP-GCN is shown in Figure~\ref{fig:framework}. It has two major components (i.e., node similarity preserving aggregation and self-supervised learning) corresponding to the two challenges, respectively. The node similarity preserving aggregation component introduces a novel adaptive strategy in Section~\ref{sec:framework1} to balance the influence between the graph structure and node features during aggregation. Then the self-supervised learning component in Section~\ref{sec:ssl} is to better preserve node similarity by considering both similar and dissimilar pairs. Next we will detail each component.

\subsection{Node Similarity Preserving Aggregation}
\label{sec:fsp}\label{sec:framework1}
In order to endow GCN with the capability of preserving feature similarity, we first propose to integrate the node features with the structure information of the original graph by constructing a new graph that can adaptively balance their influence on the aggregation process. This is achieved by first constructing a $k$-nearest-neighbor ($k$NN) graph based on the node features and then adaptively integrating it with the original graph into the aggregation process of GCN. Then, learnable self-loops are introduced to adaptively encourage the contribution of a node's own features in the aggregation process.

%
\subsubsection{Feature Graph Construction}
In feature graph construction, we convert the feature matrix ${\bf X}$ into a feature graph by generating a $k$NN graph based on the cosine similarity between the features of each node pair. For a given node pair $(v_i, v_j)$, their cosine similarity can be calculated as:
\begin{equation}\label{eq:sim}
{\bf S}_{i j}=\frac{\mathbf{x}_{i}^{\top}\mathbf{x}_{j}}{\left\|\mathbf{x}_{i}\right\|\left\|\mathbf{x}_{j}\right\|}.
\end{equation}
Then we choose $k=20$ nearest neighbors following the above cosine similarity for each node and obtain the $k$NN graph. We denote the adjacency matrix of the constructed graph as ${\bf A}_f$ and its degree matrix as ${\bf D}_f$.

\subsubsection{Adaptive Graph Integration}
The graph integration process is adaptive in two folds: (1) each node adaptively balances the information from the original  graph and the feature $k$NN graph; and (2) each node can adjust the contribution of its node features. 

After we obtain the $k$NN graph, the propagation matrix ${\bf P}^{(l)}$ in the $l$-th layer can be formulated as,
\begin{equation}
{{\bf P}^{(l)}} = {{\bf s}^{(l)}}*\tilde{\bf D}^{-1/2}\tilde{\bf A}\tilde{\bf D}^{-1/2} + (1-{\bf s}^{(l)})* {\bf D}_f^{-1/2}{\bf A}_f{{\bf D}_f^{-1/2}},
\end{equation}
 where ${{\bf s}^{(l)}} \in \mathbb{R}^{n}$ is a score vector which balances the effect of the original and feature graphs. Note that ``${\bf a} * {\bf M}$'' denotes the operation to multiply the $i$-th element vector ${\bf a}$ with the $i$-th row of the matrix ${\bf M}$. One advantage from ${{\bf s}^{(l)}}$ is that it allows nodes to adaptively combine information from these two graphs as nodes can have different scores. To reduce the number of parameters in ${{\bf s}^{(l)}} $, we model ${{\bf s}^{(l)}} $ as:
\begin{equation}
    {{\bf s}^{(l)}} = \operatorname{\sigma}\left({{\bf H}^{(l-1)} {\bf W}_s^{(l)}}+ b_s^{(l)}\right),
\label{eq:propagation1}
\end{equation}
where ${{\bf H}^{(l-1)}} \in \mathbb{R}^{n\times d^{(l-1)}}$  denotes the input hidden representation from previous layer (with ${{\bf H}^{(0)}}={\bf X}$) and ${\bf W}_s^{(l)} \in \mathbb{R}^{d^{(l-1)}\times1}$ and $b_s^{(l)}$ are the parameters for transforming ${{\bf H}^{(l-1)}}$ into the score vector ${\bf s}^{(l)}$. Here $\operatorname{\sigma}$ denotes an activation function and we apply the sigmoid function. In this way, the number of parameters for construing ${{\bf s}^{(l)}} $ has reduced from $n$ to ${d^{(l-1)}} + 1$.

\subsubsection{Adaptive Learnable Self-loops}

In the original GCN, given a node $v_i$, one self-loop is added to include its own features into the aggregation process. Thus, if we want to preserve more information from its original features, we can add more self-loops to $v_i$. However, the importance of node features could be distinct for different nodes; and different numbers of self-loops are desired for nodes. Thus, we propose to add a learnable diagonal matrix ${{\bf D}^{(l)}_K} = diag{\left(K_1^{(l)}, K_2^{(l)}, \dots, K_n^{(l)}\right)}$ to the propagation matrix ${\bf P}^{(l)}$, 
\begin{equation}
    {\bf \tilde{P}}^{(l)} = {\bf P}^{(l)} + \gamma{{\bf D}^{(l)}_K},
\end{equation}
where ${{\bf D}_K^{(l)}}$ adds learnable self-loops to the integrated graph. In particular, $K_i^{(l)}$ indicates the number of self-loops added to node $v_i$ at the $l$-th layer. $\gamma$ is a predefined hyper-parameter to control the contribution of self-loops. To reduce the number of parameters, we use a linear layer to learn $K_1^{(l)}, K_2^{(l)}, \dots, K_n^{(l)}$ as:
\begin{equation}
    {{K}_i^{(l)}} = {{\bf H}_i^{(l-1)} {\bf W}_K^{(l)}} + b_K^{(l)},
\end{equation}
where ${\bf H}_i^{(l-1)}$ is the hidden presentation of $v_i$ at the layer $l-1$; ${\bf W}_K^{(l)} \in \mathbb{R}^{d^{(l-1)}\times1}$ and $b_K^{(l)}$ are the parameters for learning the self-loops $K_1^{(l)}, K_2^{(l)}, \dots, K_n^{(l)}$.

\subsubsection{The Classification Loss}
Ultimately the hidden representations can be formulated as,
\begin{equation}
    {\bf H}^{(l)} = \operatorname{\sigma}(\tilde{\bf P}^{(l)}{\bf H}^{(l-1)}{\bf W}^{(l)}),
\end{equation}
where ${\bf W}^{(l)} \in \mathbb{R}^{d^{(l-1)} \times d^{(l)}}$, $\operatorname{\sigma}$ denotes an activation function such as the ReLU function. We denote the output of the last layer as $\hat{\bf H}$, then the classification loss is shown as,
\begin{equation}\label{eq:class}
\mathcal{L}_{class}=  \frac{1}{\vert\mathcal{D}_{L}\vert} \sum_{(v_i, y_i) \in \mathcal{D}_{L}} \ell\left(softmax(\hat{\bf H}_{i}), y_{i}\right),
\end{equation}
where $\mathcal{D}_L$ is the set of labeled nodes, $y_i$ is the label of node $v_i$ and $\ell(\cdot,\cdot)$ is the loss function to measure the difference between predictions and true labels such as cross entropy. 


\subsection{Self-Supervised Learning}
\label{sec:ssl}
 
Although the constructed $k$NN feature graph component in Eq. ~(\ref{eq:propagation1}) plays a critical role of pushing nodes with similar features to become similar, it does not directly model feature dissimilarity. In other words, it only can effectively preserve the top similar pairs of nodes to have similar representations and will not push nodes with dissimilar original features away from each other in the learned embedding space. Thus, to better preserve pairwise node similarity, we incorporate the complex pairwise node relations by proposing a contrastive self-supervised learning component to SimP-GCN. 


Recently, self-supervised learning techniques have been applied to graph neural networks for leveraging the rich information in tremendous unlabeled nodes~\cite{hassani2020contrastive, you2020does,jin2020selfsupervised,you2020graph}. Self-supervised learning first designs a domain specific pretext task to assign constructed labels for nodes and then trains the  model on the pretext task to learn better node representations. Following the joint training manner described in~\cite{jin2020selfsupervised, you2020does}, we design a contrastive pretext task where the self-supervised component is asked to predict pairwise feature similarity. In detail, we first calculate pairwise similarity for each node pair and sample node pairs to generate self-supervised training samples. Specifically, for each node, we sample its $m$ most similar nodes and $m$ most dissimilar nodes. Then the pretext task can be formulated as a regression problem and the self-supervised loss can be stated as,
\begin{equation}\label{eq:self}
\mathcal{L}_{self}( {\bf A}, {\bf X}) =  \frac{1}{\vert\mathcal{T}\vert}\sum_{(v_i,v_j) \in \mathcal{T}} \| f_{w} ({\bf H}_{i}^{(l)} - {\bf H}_{j}^{(l)}) - {{\bf S}_{ij}} \|^2,
\end{equation}
where $\mathcal{T}$ is the set of sampled node pairs, $f_{w}$ is a linear mapping function, ${\bf S}_{ij}$ is defined in Eq.~(\ref{eq:sim}), and ${\bf H}_{i}^{(l)}$ is the hidden representation of node $v_i$ at $l$-th layer. Note that we can formulate the pretext task as a classification problem to predict whether the given node pair is similar or not (or even the levels of similarity/dissimilarity). However, in this work we adopt the regression setting and use the first layer's hidden representation, i.e., $l = 1$ while setting $m = 5$.

\subsection{Objective Function and Complexity Analysis}
With the major components of SimP-GCN, next we first present the overall objective function where we jointly optimize the classification loss along with the self-supervised loss. Thereafter, we present a complexity analysis and discussion on the proposed framework.

\subsubsection{Overall Objective Function}
The overall objective function can be stated as,
\begin{align}
\min \mathcal{L} &= \mathcal{L}_{class}  + \lambda\mathcal{L}_{self} \\ 
&= \frac{1}{\vert\mathcal{D}_{L}\vert} \sum_{(v_i, y_i) \in \mathcal{D}_{L}} \ell\left(softmax(\hat{\bf H}_{i}), y_{i}\right) \nonumber \\
& \hspace{2.5ex}+ \frac{\lambda}{\vert\mathcal{T}\vert}\sum_{(v_i,v_j) \in \mathcal{T}} \| f_{w} ({\bf H}_{i}^{(1)} - {\bf H}_{j}^{(1)}) - {{\bf S}_{ij}} \|^2 \nonumber
\end{align}
where $\lambda$ is a hyper-parameter that controls the contribution of self-supervised loss (i.e., $\mathcal{L}_{self}$ of Eq.~(\ref{eq:self})) in addition to the traditional classification loss (i.e., $\mathcal{L}_{class}$ of Eq.~(\ref{eq:class})). 
\subsubsection{Complexity Analysis}
Here, we compare the proposed method, SimP-GCN, with vanilla GCN by analyzing the additional complexity in terms of time and model parameters.  

\vskip 0.5em
\noindent{}\textbf{Time Complexity.}
In comparison to vanilla GCN, the additional computational requirement mostly comes from calculating the pairwise feature similarity for the $k$NN graph construction and self-supervised component. Its time complexity is $O(n^2)$ 
if done na\"ively. However, calculating pariwise similarity can be made significantly more efficient. For example, the calculations are inherently parallelizable. In addition, this has been a well studied problem and in the literature there exist approximation methods that could be used to significantly speedup the computation for larger graphs, such as~\cite{chen2009fast} that presented a divide-and-conquer method via Recursive Lanczos Bisection, or~\cite{dong2011efficient} that is inherently suitable for MapReduce where they empirically achieved approximate $k$NN graphs in $O(n^{1.14})$.

\vskip 0.5em
\noindent{}\textbf{Model Complexity.}
As shown in Section~\ref{sec:fsp} and Section~\ref{sec:ssl}, our model introduces additional parameters ${\bf W}_s^{(l)}$, $b_s^{(l)}$, ${\bf W}_K^{(l)}$, $b_K^{(l)}$ and linear mapping $f_w$. It should be noted that ${\bf W}_s^{(l)}, {\bf W}_K^{(l)}\in \mathbb{R}^{d^{(l-1)}\times1}$ and $b_s^{(l)}, b_K^{(l)} \in \mathbb{R}$. Since $f_w$ transforms $\bf{H}$ to a one-dimensional vector, the weight matrix in $f_w$ has a shape of $d^{(l)}\times 1$. Hence, compared with GCN, the overall additional parameters of our model are $O(d^{(l)})$ where $d^{(l)}$ is the input feature dimension in the $l$-th layer. It suggests that our model only introduces additional parameters that is linear to the feature dimension.

%% file: section/experiment.tex

\begin{table}[t]\small
\caption{Dataset Statistics.}
\vspace{-1ex}
\begin{tabular}{@{}lccc|cccc@{}}
\toprule
           & \multicolumn{3}{c}{\textbf{Assortative}} & \multicolumn{4}{c}{\textbf{Disassortative}}  \\ 
\textbf{Datasets}   & Cora    & Cite.  & Pubm.   & Actor & Corn. & Texas & Wisc. \\\midrule
\#Nodes    & 2,708    & 3,327       & 19,717    & 7,600  & 183     & 183   & 251       \\
\#Edges    & 5,429    & 4,732       & 44,338    & 33,544 & 295     & 309   & 499       \\
\#Features & 1,433    & 3,703       & 500      & 931   & 1,703    & 1,703  & 1,703      \\
\#Classes  & 7       & 6          & 3        & 5     & 5       & 5     & 5         \\ \bottomrule
\end{tabular}
\vspace{-1ex}
\label{tab:dataset}
\end{table}

\section{Experiment}
In this section, we evaluate the effectiveness of SimP-GCN under different settings. In particular, we aim to answer the following questions: 

\begin{itemize}[leftmargin=*]
    \item \textbf{RQ1:} How does SimP-GCN perform on both assortative and disassortative graphs?
    \item \textbf{RQ2:} Can SimP-GCN boost model robustness and defend against graph adversarial attacks?
    \item \textbf{RQ3:} Does the framework work as designed? and How do different components affect the performance of SimP-GCN? 
\end{itemize}
Note that though there are many scenarios where node similarity is important, we focus on the scenarios of disassortative graphs and graphs manipulated by adversarial attacks to illustrate the advantages of the proposed framework by preserving node similarity.

\subsection{Experimental Settings}
\subsubsection{Datasets.} Since the performance of graph neural networks can be significantly different on assortative and disassortative graphs, we select several representative datasets from both categories to conduct the experiments. Specifically, for assortative graphs we adopt three citation networks that are commonly used in the GNN literature, i.e., Cora, Citeseer and Pubmed~\cite{kipf2016semi}. For disassortative graphs, we use one actor co-occurrence network, i.e., Actor~\cite{tang2009social}, and three webpage datasets, i.e., Cornell, Texas and Wisconsin~\cite{pei2020geom}. The statistics of these datasets are shown in Table~\ref{tab:dataset}.

\subsubsection{Baselines}
\label{sec:baselines}
To evaluate the effectiveness of the proposed framework, we choose the following representative semi-supervised learning baselines including the state-of-the-art GNN models:
\begin{itemize}[leftmargin=*]
    \item \textbf{LP}~\cite{zhu2003semi-lp}: Label Propagation (LP) explores structure and label information by a Gaussian random field model. Note that LP does not exploit node features. 
    \item \textbf{GCN}~\cite{kipf2016semi}: GCN is one of the most popular graph convolutional models and our proposed model is based on it. 
    \item \textbf{$k$NN-GCN}~\cite{franceschi2019learning-discrete}: It is a variant of GCN which takes the $k$-nearest neighbor graph created from the node features as input. Here we use it as a baseline to check the performance of directly employing the $k$NN graph without using the original graph.
    \item \textbf{(${\bf A}$+$k$NN)-GCN}: GCN that takes ${\bf A} + {\bf A}_f$ as the input graph. 
    \item \textbf{GAT}~\cite{gat}: Graph attention network (GAT) employs attention mechanism and learns different scores to different nodes within the neighborhood. It is widely used as a GNN baseline.
    \item \textbf{JK-Net}~\cite{xu2018representation}: JK-Net uses dense connections to leverage different neighbor ranges for each node to learn better representations. 
    \item  \textbf{GCNII}~\cite{chen2020simple}: Based on GCN, GCNII employs residual connection and identity mapping to achieve better performance. GCNII* is a variant of GCNII with more parameters. 
    \item  \textbf{Geom-GCN}~\cite{pei2020geom}: Geom-GCN explores to capture long-range dependencies in disassortative graphs. It  uses the geometric relationships defined in the latent space to build structural neighborhoods for aggregation. There are three variants of Geom-GCN: Geom-GCN-I, Geom-GCN-P and Geom-GCN-S. 
\end{itemize}
Note that \textbf{Geom-GCN} is mainly designed for disassortative graphs; thus we only report its performance on disassortative graphs.

\subsubsection{Parameter Setting}  
For experiments on assortative graphs, we set the number of layers in GCN and SimP-GCN to 2,  with hidden units 128, $L_2$ regularization $5e{-4}$, dropout rate $0.5$, epochs $200$ and learning rate $0.01$. In SimP-GCN, the weighting parameter $\lambda$ is searched from $\{0.1, 0.5, 1, 5, 10, 50, 100\}$, $\gamma$ is searched from $\{0.01, 0.1\}$ and the initialization of $b_s$ is searched from $\{0, 2\}$. Since GCNII and JK-Net with multi-layers can use much higher memory due to more parameters, we restrict the depth to $4$. For experiments on disassortative graphs, we set the number of layers in GCN and SimP-GCN to 2, with learning rate 0.05, dropout rate 0.5, epochs 500 and patience 100. The number of hidden units is tuned from \{16, 32, 48\} and $L_2$ regularization is tuned from $\{$5$e$-4, 5$e$-5$\}$. SimP-GCN further searches $\lambda$ from \{0.1, 1, 10\} and $\gamma$ from \{0.01, 0.1, 1\}. For other baseline methods, we use the default parameter settings in the authors’ implementation.



\subsection{Performance Comparison}
In this subsection, we answer the first question and compare the performance on assortative and disassortative graphs, respectively. 

\subsubsection{Performance Comparison on Assortative Graphs.}
For the experiments on assortative graphs, we follow the widely used semi-supervised setting in~\cite{kipf2016semi} with 20 nodes per class for training, 500 nodes for validation and 1000 nodes for testing. We report the average accuracy of 10 runs in Table~\ref{tab:assortative}. On the three assortative graphs, SimP-GCN consistently improves GCN and achieves the best performance in most settings. The improvement made by SimP-GCN demonstrates that preserving feature similarity can benefit GCN in exploiting structure and feature information. It is worthwhile to note that LP and $k$NN, which only take the graph structure or node features as input, can achieve reasonable performance, but they cannot beat GCN. The variant (${\bf A}$+$k$NN)-GCN cannot outperform $k$NN-GCN or GCN. This suggests that learning a balance between graph structure and node features is of great significance.

\begin{table}[t]
\caption{Node classification accuracy (\%) on assortative graphs. The best performance is highlighted in bold.}
\vskip -0.25em
\begin{tabular}{@{}lccc@{}}
\toprule
\textbf{Method}  & \multicolumn{1}{l}{\textbf{Cora}} & \multicolumn{1}{l}{\textbf{Citeseer}} & \multicolumn{1}{l}{\textbf{Pubmed}} \\ \midrule
LP      & 74.6                     & 57.7                         & 71.6                       \\
GCN     & 81.3                     & 71.5                         & 79.3                       \\
$k$NN-GCN & 67.4                     & 68.7                         & 78.9                       \\
({\bf A}+$k$NN)-GCN &79.1 & 71.1&  80.8 \\
GAT     & \textbf{83.1}            & 70.8                         & 78.5                       \\
JK-Net   & 80.3                     & 68.5                         & 78.3                       \\
GCNII   & 82.6                     & 68.9                         & 78.8                       \\
GCNII*  & 82.3                     & 67.9                         & 78.2                       \\ \midrule
SimP-GCN    & 82.8                     & \textbf{72.6}                & \textbf{81.1}              \\ \bottomrule
\end{tabular}
\vskip -0.25em
\label{tab:assortative}
\end{table}

\begin{table}[h]
\caption{Node classification accuracy (\%) on disassortative graphs. The best performance is highlighted in bold.}
\vskip -0.25em
\begin{tabular}{@{}lcccc@{}}
\toprule
\textbf{Method}    & \multicolumn{1}{l}{\textbf{Actor}} & \multicolumn{1}{l}{\textbf{Cornell}} & \multicolumn{1}{l}{\textbf{Texas}} & \multicolumn{1}{l}{\textbf{Wisconsin}} \\ \midrule
LP         & 23.53                     & 35.95                       & 32.70                      & 29.41                         \\
GCN        & 26.86                     & 52.70                        & 52.16                     & 45.88                         \\
$k$NN-GCN    & 33.77                     & 81.35                       & 79.73                     & 81.35                           \\ 
({\bf A}+$k$NN)-GCN    & 32.83 & 76.49 & 79.60  & 80.98  \\
GAT        & 28.45                     & 54.32                       & 58.38                     & 49.41                         \\
JK-Net      & 27.41                      & 57.84                       & 55.95                     & 50.78                         \\
Geom-GCN-I & 29.09                     & 56.76                       & 57.58                     & 58.24                         \\
Geom-GCN-P & 31.63                     & 60.81                       & 67.57                     & 64.12                         \\
Geom-GCN-S & 30.30                      & 55.68                       & 59.73                     & 56.67                         \\
GCNII      & 33.91                     & 74.86                       & 69.46                     & 74.12                         \\
GCNII*     & 35.18                     & 76.49                       & 77.84                     & 81.57                         \\ \midrule
SimP-GCN      & \textbf{36.20}             & \textbf{84.05}              & \textbf{81.62}            & \textbf{85.49}                \\ \bottomrule
\end{tabular}
\vskip -0.25em
\label{tab:disassortative}
\end{table}

\begin{figure*}[!tb]%
     \centering
     \subfloat[Cora]{{\includegraphics[width=0.33\linewidth]{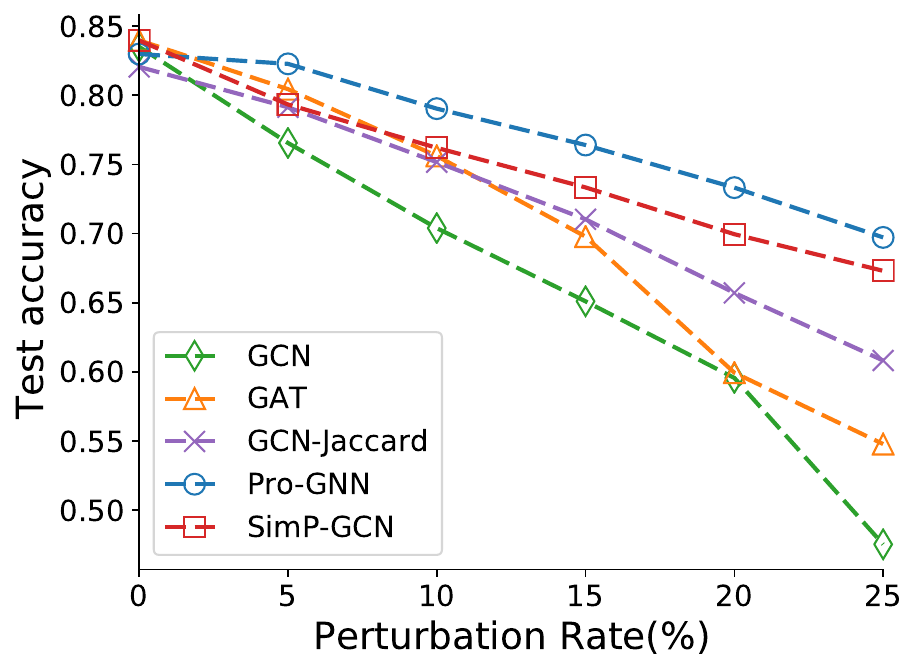} }\label{fig:param-lambda}}%
     \subfloat[Citeseer]{{\includegraphics[width=0.33\linewidth]{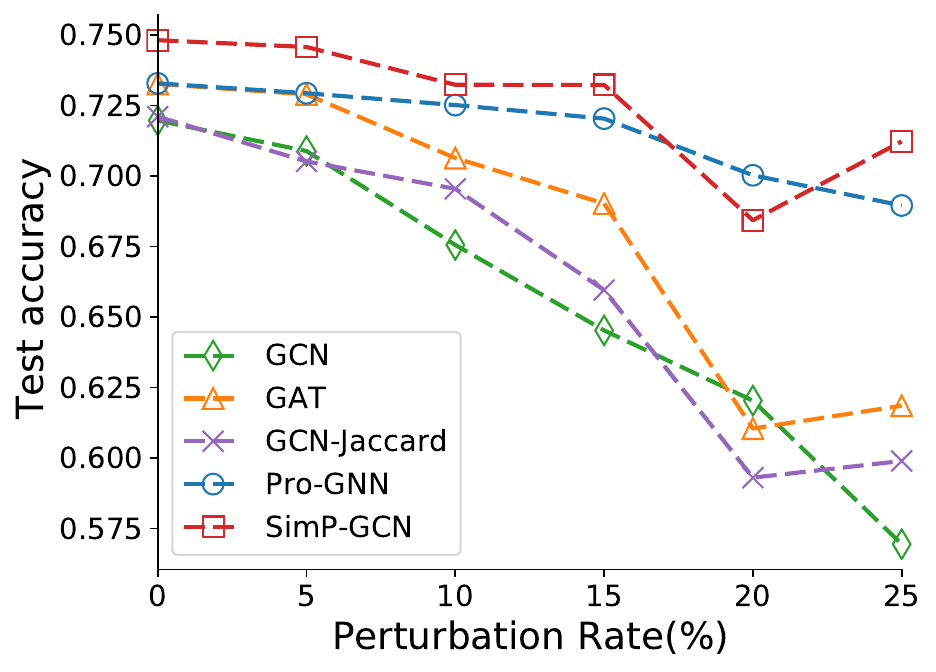} }\label{fig:param-alpha}}%
    \subfloat[Pubmed]{{\includegraphics[width=0.33\linewidth]{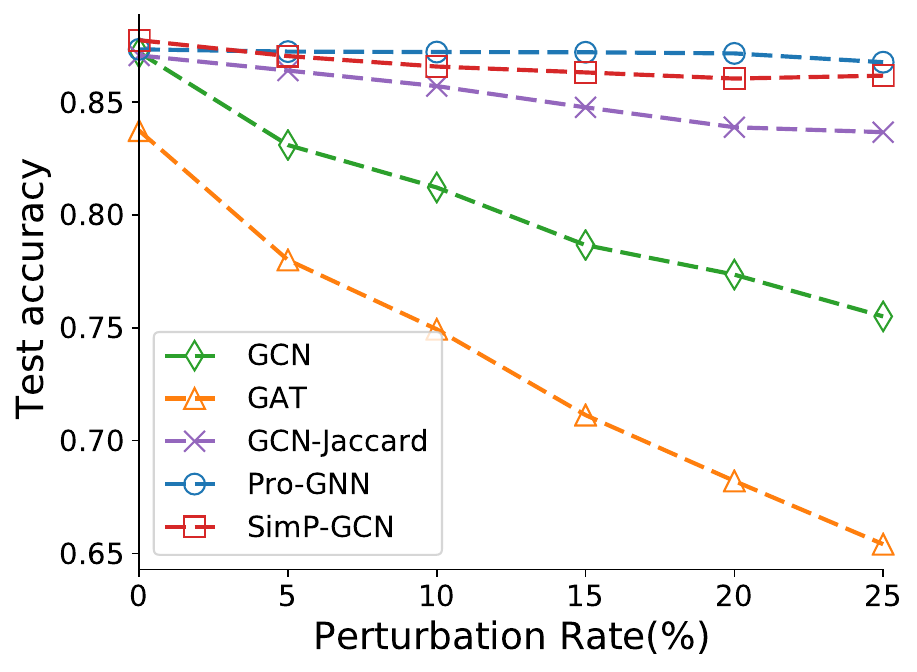} }\label{fig:param-alpha}}%
    \qquad
    \vskip -0.5em
    \caption{Node classification accuracy under non-targeted attack (\textit{metattack}).} 
    \label{fig:robustness}
\vskip -0.75em
\end{figure*}

\subsubsection{Performance Comparison on Disassortative Graphs.}
\label{sec:disassorative}
In this experiment, we report the performance on four disassortative graphs, i.e., Actor, Cornell, Texas and Wisconsin. Following the common setting of experiments on disassortative graphs~\cite{pei2020geom,liu2020non}, we randomly split nodes of each class into 60\%, 20\%, and 20\% for training, validation and test. 

Note that both Geom-GCN and GCNII are the recent models that achieve the state-of-the-art performance on disassortative graphs. We report the average accuracy of all models on the test sets over 10 random splits in Table~\ref{tab:disassortative}.
We reuse the metrics reported in~\cite{chen2020simple} for GCN, GAT and Geom-GCN. From the results we can see that the performance of LP is extremely low on all datasets, which indicates that structure information is not very useful for the downstream task. Besides, the fact that GCN, GAT, JK-Net and Geom-GCN cannot outperform $k$NN-GCN verifies that feature information in those disassortative graphs are much more important and the original structure information could even be harmful. We note that (${\bf A}$+$k$NN)-GCN cannot improve $k$NN-GCN. It supports that simply combining structure and features is not sufficient. SimP-GCN consistently improves GCN by a large margin and achieves state-of-the-art results on four disassortative graphs.

\begin{figure*}[t]%
\vskip -0.2em
     \centering
     \subfloat[Assortative graphs]{{\includegraphics[width=0.49\linewidth]{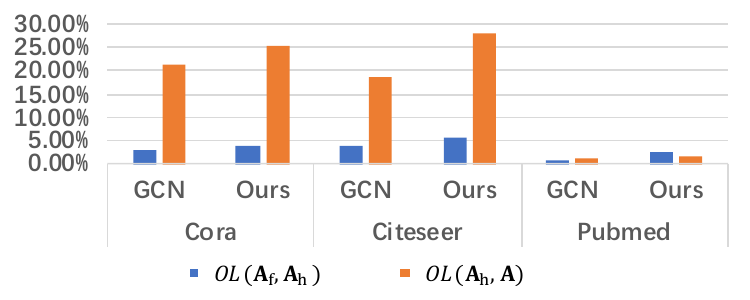} }\label{fig:param-lambda}}%
     \subfloat[Disassortative graphs]{{\includegraphics[width=0.49\linewidth]{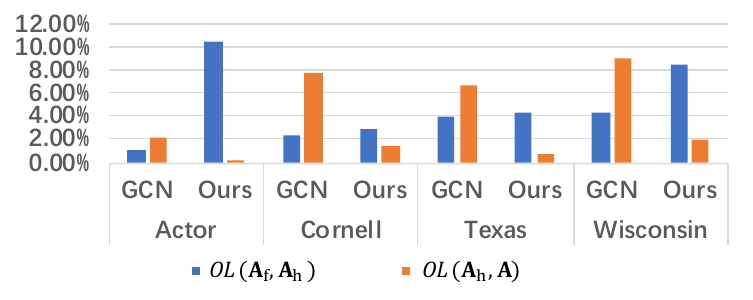} }\label{fig:param-alpha}}%
    \qquad
    \vskip -0.5em    
    \caption{Overlapping percentage on assortative and disassortative graphs. ``Ours" means SimP-GCN. Blue bar indicates the overlapping between feature and hidden graphs; and orange bar denotes the overlapping between hidden and original graphs.} 
\vskip -0.5em
\label{fig:overlapping}%
\end{figure*}

\subsection{Adversarial Robustness}
Traditional deep neural networks can be easily fooled by adversarial attacks~\cite{xu2019adversarial, li2020learning,li2019regional}. Similarly, recent research has also demonstrated that graph neural networks are vulnerable to adversarial attacks~\cite{nettack, jin2020graph,sun2020adversarial}. In other words, unnoticeable perturbation on graph structure can significantly reduce their performances. Such vulnerability has raised great concern for applying graph neural networks in safety-critical applications. Although attackers can perturb both node features and graph structures, the majority of existing attacks focus on changing graph structures. Recall that SimP-GCN is shown to greatly improve GCN when the structure information is not very useful in Section~\ref{sec:disassorative}. Therefore, in this subsection, we are interested in examining its potential benefit on adversarial robustness. Specifically, we evaluate the node classification performance of SimP-GCN under non-targeted graph adversarial attacks. The goal of non-targeted attack is to degrade the overall performance of GNNs on the whole node set. We adopt \textit{metattack}~\cite{mettack} as the attack method. We focus on attacking graph structure and vary the perturbation rate, i.e., the ratio of changed edges, from 0 to 25\% with a step of 5\%. To make better comparison, we include GCN, GAT and the state-of-the-art defense models, GCN-Jaccard~\cite{wu2019adversarial-jaccard} and  Pro-GNN~\cite{jin2020graph} implemented by DeepRobust\footnote{\url{https://github.com/DSE-MSU/DeepRobust}}~\cite{li2020deeprobust}, as baselines and use the default hyper-parameter settings in the authors' implementations. For each dataset, the hyper-parameters of SimP-GCN are set as the same values under different perturbation rates. 
We follow the parameter settings of baseline and attack methods in~\cite{jin2020graph} and evaluate models' robustness on Cora, Citeseer and Pubmed datasets. Furthermore, as suggested in ~\cite{jin2020graph}, we randomly choose 10\% of nodes for training, 10\% of nodes for validation and 80\% of nodes for test. All experiments are repeated 10 times and the average accuracy of different methods under various settings is shown in Figure~\ref{fig:robustness}. 

As we can observe from Figure~\ref{fig:robustness}, SimP-GCN consistently improves the performance of GCN under different perturbation rates of adversarial attack on all three datasets. Specifically, SimP-GCN improves GCN by a larger margin when the perturbation rate is higher. For example, it achieves over 20\% improvement over GCN under the 25\% perturbation rate on Cora dataset. This is because the structure information will be very noisy and misleading when the graph is heavily poisoned. In this case, node feature similarity can provide strong guidance to train the model and boost model robustness. SimP-GCN always outperforms GCN-Jaccard and shows comparable performance to Pro-GNN on all datasets. It even obtains the best performance on the Citeseer dataset. These observations suggest that by preserving node similarity, SimP-GCN can be robust to existing adversarial attacks.  

\begin{figure*}[t]%
\vskip -0.4em
     \centering
     \subfloat[Cora]{{\includegraphics[width=0.33\linewidth]{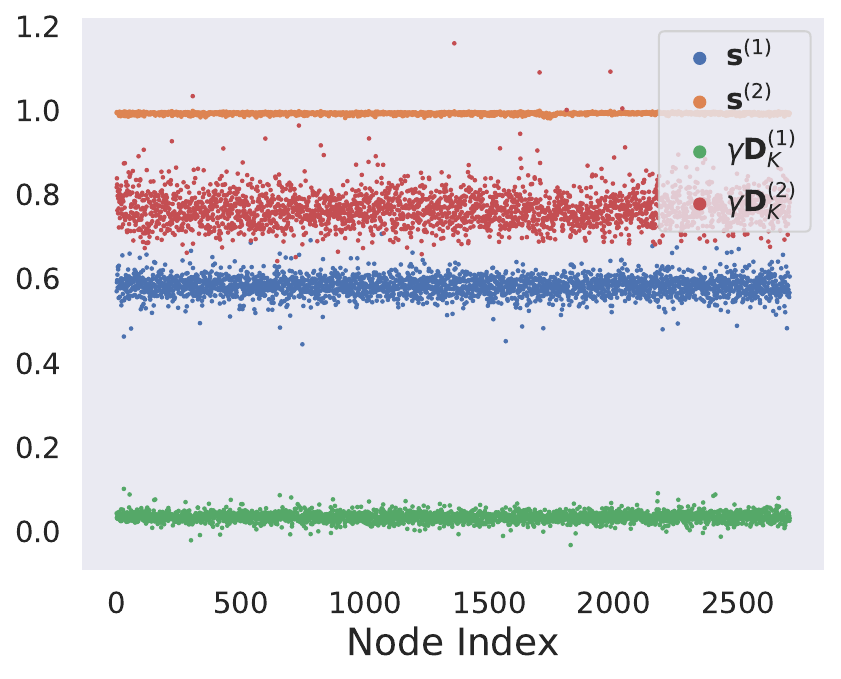} }}%
     \subfloat[Actor]{{\includegraphics[width=0.33\linewidth]{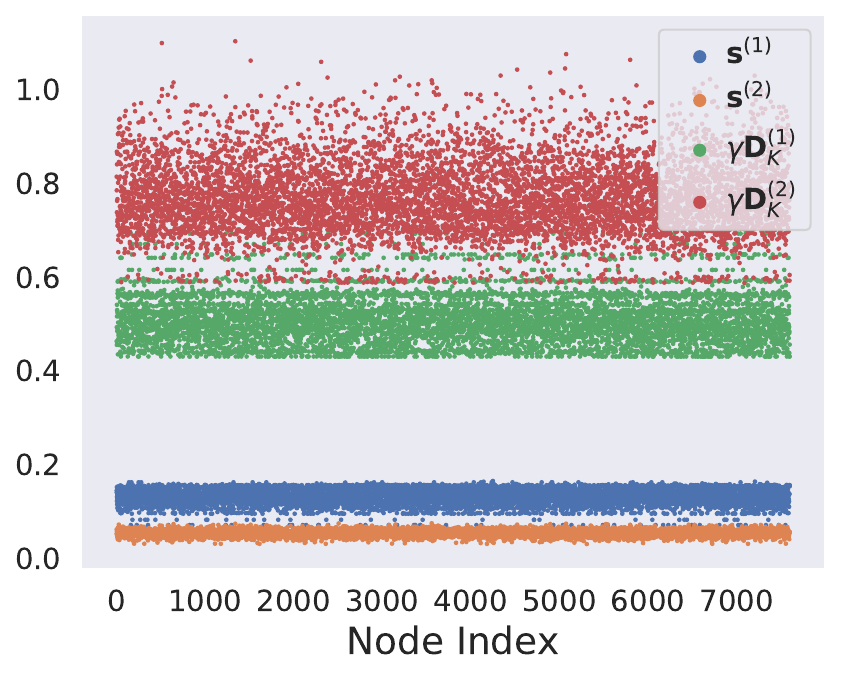} }}%
     \subfloat[Wisconsin]{{\includegraphics[width=0.33\linewidth]{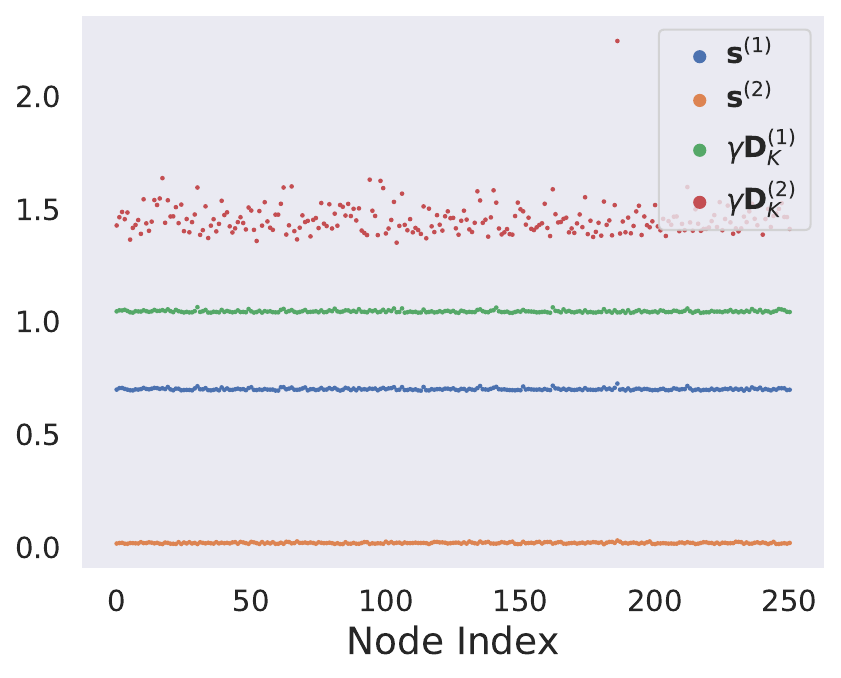} }}%
    \qquad
\vskip -0.5em
\caption{Learned values of ${\bf s}^{(1)}$, ${\bf s}^{(2)}$, ${\bf D}_n^{(1)}$ and ${\bf D}_n^{(2)}$ on Cora, Actor and Wisconsin.} 
\vskip -1em
\label{fig:scores}%
\end{figure*}


\subsection{Further Probe}
In this subsection, we take a deeper examination on the proposed framework to understand how it works and how each component affects its performance. 

\subsubsection{Is Node Similarity Well Preserved?} Next, we investigate whether SimP-GCN can preserve node feature similarity. Following the same evaluation strategy in Section~\ref{sec:overlapping}, we focus on the overlapping percentage of graphs pairs. Specifically, we construct $k$NN graphs from the learned hidden representations of GCN and SimP-GCN, denoted as ${\bf A}_h$, and calculate the overlapping percentages $OL({\bf A}_f, {\bf A}_h)$ and $OL({\bf A}_h, {\bf A})$ on all datasets. The results are shown in Figure~\ref{fig:overlapping}. From the results, we make the following observations:
\begin{itemize}[leftmargin=*]
\item For both assortative and disassortative graphs, SimP-GCN always improves the overlapping percentage between feature and hidden graphs, i.e., $OL({\bf A}_f, {\bf A}_h)$. It shows that SimP-GCN can effectively preserve more information from feature similarity. Especially in disassortative graphs where homophily is low, SimP-GCN essentially boosts $OL({\bf A}_f, {\bf A}_h)$ by a large margin, e.g., roughly 10\% on the Actor dataset. 
\item In disassortative graphs, SimP-GCN decreases the overlapping percentage between hidden and original graphs, i.e., $OL({\bf A}_h, {\bf A})$. One reason is that the structure information in disassortative graphs is less useful (or even harmful) to the downstream task. Such phenomenon is also in line with our observation in Section~\ref{sec:disassorative}. On the contrary, in assortative graphs both $OL({\bf A}_h, {\bf A})$ and $OL({\bf A}_f, {\bf A}_h)$ are improved, which indicates that SimP-GCN can explore more information from both structure and features for assortative graphs. 
\end{itemize}

\subsubsection{Do the Score Vectors and Self-loop Matrices Work?} To study how the score vectors $({\bf s}^{(1)}, {\bf s}^{(2)})$ and learnable self-loop matrices $({\bf D}_n^{(1)}, {\bf D}_n^{(2)})$ in Eq. (\ref{eq:propagation1}) benefit SimP-GCN, we set $\gamma$ to $0.1$ and visualize the learned values in Figure~\ref{fig:scores}. Due to the page limit, we only report the results on Cora, Actor and Wisconsin. From the results, we have the following findings: 
\begin{itemize}[leftmargin=*]
    \item The learned values vary in a range, indicating that the designed aggregation scheme can adapt the information differently for individual nodes. For example, in Cora dataset, ${\bf s}^{(1)}$, ${\bf s}^{(2)}$, ${\gamma\bf D}_n^{(1)}$ and ${\gamma\bf D}_n^{(2)}$ are in the range of $[0.44, 0.71]$, $[0.98, 1.00]$, $[-0.03, 0.1]$ and $[0.64, 1.16]$, respectively. 
    \item On the assortative graph (Cora), $\gamma{\bf D}_n^{(1)}$ is extremely small and the model is mainly balancing the information from original and feature graphs. On the contrary, in disassortative graphs (Actor and Wisconsin),  ${\gamma\bf D}_n^{(1)}$ has much larger values, indicating that node original features play a significant role in making predictions for disassortative graphs.
\end{itemize}

\subsubsection{Ablation Study}
To get a better understanding of how different components affect the model performance, we conduct ablation studies and answer the third question in this subsection. Specifically, we build the following ablations:
\begin{itemize}[leftmargin=*]
    \item Keeping self-supervised learning (SSL) component and all other components. This one is our original proposed model.
    \item Keeping SSL component but removing the learnable diagonal matrix ${\bf D}_n$ from the propagation matrix ${\bf \tilde{P}}$, i.e., setting $\gamma$ to 0.
    \item Removing SSL component.
    \item Removing SSL component and the learnable diagonal matrix ${\bf D}_n$. 
\end{itemize}
Since SimP-GCN achieves the most significant improvement on disassortative graphs, we only report the performance on disassortative graphs. We use the best performing
hyper-parameters found for the results in Table~\ref{tab:disassortative} and report the average accuracy of 10 runs in Table~\ref{tab:ablation}. By comparing the ablations with GCN, we observe that all components contribute to the performance gain: ${\bf A}_f$ and ${\bf D}_n$ essentially boost the performance while the SSL component can further improve the performance based on ${\bf A}_f$ and ${\bf D}_n$.

\begin{table}[t]\small
\caption{Ablation study results (\% test accuracy) on disassortative graphs.}
\vskip -0.9em
\begin{tabular}{@{}lcccc@{}}
\toprule
\textbf{Ablation}                                                          & \textbf{Actor} & \textbf{Corn.} & \textbf{Texa.} & \textbf{Wisc.} \\ \midrule
SimP-GCN & & & & \\
- with SSL and  $({\bf A}, {\bf A}_f, {\bf D}_n)$                                              & 36.20          & 84.05            & 81.62          & 85.49              \\
- with SSL and  $({\bf A}, {\bf A}_f)$                                                    & 35.75          & 65.95            & 71.62          & 81.37              \\
- w/o SSL and with $({\bf A}, {\bf A}_f,{\bf D}_n)$ & 36.09          & 82.70            & 79.73          & 84.12              \\
- w/o SSL and with $({\bf A}, {\bf A}_f)$                                                 & 34.68          & 64.59            & 68.92          & 81.57              \\ \midrule
GCN        & 26.86                     & 52.70                        & 52.16                     & 45.88                         \\
\bottomrule
\end{tabular}
\label{tab:ablation}
\vskip -0.75em
\end{table}

%% file: section/conclusion.tex
\section{Conclusion}
Graph neural networks extract effective node representations by aggregating and transforming node features within the neighborhood. We show that the aggregation process inevitably breaks node similarity of the original feature space through theoretical and empirical analysis. Based on these findings, we introduce a node similarity preserving graph convolutional neural network model, SimP-GCN, which effectively and adaptively balances the structure and feature information as well as captures pairwise node similarity via self-supervised learning. Extensive experiments demonstrate that SimP-GCN outperforms representative baselines on a wide range of real-world datasets. As one future work, we plan to explore the potential of exploring node similarity in other scenarios such as the over-smoothing issue and low-degree nodes. 

\section{Acknowledgements}

Wei Jin, Tyler Derr, Yiqi Wang, Yao Ma and Jiliang Tang are supported by the National Science Foundation of the United States under CNS1815636, IIS1928278, IIS1714741, IIS1845081, IIS1907704 and IIS1955285. Zitao Liu is supported by the Beijing Nova Program (Z201100006820068) from Beijing Municipal Science \& Technology Commission.